# Optimizing Genetic Algorithms with Multilayer Perceptron Networks for Enhancing TinyFace Recognition

## Optimización de algoritmos genéticos con redes de perceptrones multicapa para mejorar el reconocimiento de rostros diminutos

Mohammad Subhi Al-Batah[1]  , Mowafaq Salem Alzboon[1]  , Muhyeeddin Alqaraleh[2]

[1]Jadara University, Faculty of Information Technology. Irbid, Jordan.
[2]Zarqa University, Faculty of Information Technology. Zarqa, Jordan.

**ABSTRACT**

This study conducts an empirical examination of MLP networks investigated through a rigorous methodical experimentation process involving three diverse datasets: TinyFace, Heart Disease, and Iris. Study Overview: The study includes three key methods: a) a baseline training using the default settings for the Multi-Layer Perceptron (MLP), b) feature selection using Genetic Algorithm (GA) based refinement c) Principal Component Analysis (PCA) based dimension reduction. The results show important information on how such techniques affect performance. While PCA had showed benefits in low-dimensional and noise-free datasets GA consistently increased accuracy in complex datasets by accurately identifying critical features. Comparison reveals that feature selection and dimensionality reduction play interdependent roles in enhancing MLP performance. The study contributes to the literature on feature engineering and neural network parameter optimization, offering practical guidelines for a wide range of machine learning tasks.

**RESUMEN**

Este estudio investiga la optimización de las redes de perceptrones multicapa (MLP) a través de un marco de experimentación sistemático aplicado a tres conjuntos de datos distintos: TinyFace, Heart Disease e Iris. La investigación abarca tres metodologías clave: entrenamiento de referencia con parámetros MLP predeterminados, refinamiento a través de la selección de características impulsada por algoritmos genéticos (GA) y reducción de dimensionalidad mediante análisis de componentes principales (PCA). Los hallazgos revelan información importante sobre el impacto de estas técnicas en el rendimiento del modelo. GA mejoró de manera constante la precisión en conjuntos de datos complejos al identificar de manera efectiva características críticas, mientras que PCA demostró su utilidad en conjuntos de datos de baja dimensión y sin ruido. El análisis comparativo destaca la interacción matizada entre la selección de características y la reducción de dimensionalidad, enfatizando sus roles complementarios en la optimización del rendimiento de MLP. Este estudio avanza el discurso sobre la ingeniería de características y la optimización de redes neuronales, brindando información útil para diversas aplicaciones de aprendizaje automático.

**Palabras clave:** Reconocimiento de Caras Pequeñas; Perceptrón Multicapa (MLP); Algoritmos Genéticos (GA); Análisis De Componentes Principales (PCA); Reducción de Dimensionalidad; Optimización del Aprendizaje Automático.

## INTRODUCTION

Optimal neural network performance has emerged with critical importance in the rapidly evolving domain of artificial intelligence (AI) and machine learning (ML). Multilayer Perceptron (MLP) networks represent one of the primary architectures in supervised learning, which is crucial for solving classification and prediction problems in a variety of different application areas.[1,2,3] Yet, while working with datasets is small-sized and less complex, using standard training methodologies leads to a great performance concerning efficiency, accuracy, and interpretability. This has raised the shift toward using more advanced methods for both feature selection and dimensionality reduction to improve the model performance while reducing the computational cost.

Feature selection is a crucial preprocessing phase that helps to identify and keep only the most significant features [4,5,6] reducing noise and enabling better generalization of the models. In the same spirit, dimensionality reduction techniques like PCA address the curse of dimensionality by re-representing data in lower-dimensional space while still retaining its variance.[7,8,9] These techniques are especially useful for high-dimensional and noisy datasets such as the TinyFace dataset, a state-of-the-art benchmark for low-resolution face recognition tasks.[10,11,12]

We map and discuss the relationship between feature selection, dimensionality reduction and neural network optimisation and discuss both GA and PCA as primary techniques used in each.[13,14,15] GA leverages principles of evolution to find the optimal feature set, while PCA offers a mathematical formula for generating its principal components. Selected techniques are applied to three datasets, TinyFace, Heart Disease and Iris, covering a wide range of complexity, dimensionality and domain specificity. This study aims to shed light on this issue and provide a better understanding of how MLP performance can be maximized by analyzing the effect of these techniques on their performance.

This study offers three specific contributions: (1) showing that GA is effective in improving on the feature selection of systems operating on different datasets with varied characteristics, (2) assessing how well PCA works with MLP matrices to enhance performance with MLP performance with dimensionality challenges, and (3) conducting a comparative study of feature selection and dimensionality reduction methods, further partaking in ongoing dialogue on neural network optimization strategies.[16,17,18] These results further our knowledge of these techniques, but also provide practical insights on their application on a myriad of machine learning problems.

Feature selection is one of the basic machine learning steps since it affects model interpretability, efficiency, and accuracy significantly. Traditional methods (filter and wrapper approaches) are based on statistical measures or exhaustive search techniques to select relevant features. However, these approaches often fail for high-dimensional datasets; they are computationally expensive and prone to overfitting.[19,20,21] Feature selection can also be achieved through evolutionary algorithms, and a prominent example is GA, which utilizes the principles of natural selection to develop the optimal subset of features by preserving strong combinations of features.[22,23,24] For instance, in various domains, including biomedical diagnostics and image classification, GA has been shown to improve model performance.[25,26,27]

This has proven to be particularly effective in neural networks, where the GA-based feature selection has produced steady results. As an example,[28,29,30] showed that GA could improve the accuracy of MLP networks in diagnosing heart disease by detecting important features. In the same fashion,[31,32,33] proposed using GA for image recognition problems, obtaining great reduction in training time and higher accuracy for the classifiers.

Dimensionality reduction techniques are essential for handling high-dimensional datasets as they can eliminate noise and redundancy which often negatively affects the performance of models. Principal Component Analysis (PCA) is among the most popular methods, which seeks to reduce dimensionality by projecting the data to a lower-dimensional space while retaining components with high variance.[34,35,36] Related Article: PCA is exceptionally computational efficient but applicable mostly towards datasets which are inherently linear. However, it has been surprisingly successful in areas as diverse as genomics and image processing.[37,38,39] As an illustrations,[40,41,42] implemented PCA to minimize datasets space in facial recognition, giving remarkable enhancement in classification time with no decrease in precision.

There has been great attention for the integration of feature selection and dimension reduction with neural networks training. These preprocessing techniques are essential in addressing the unique challenges posed by the low-resolution facial images in the TinyFace dataset. Multiple studies have underlined the performance achieved by using PCA along with higher-order architectures such as VGG19 for feature extraction on low-quality image datasets.[43,44,45] Likewise, studies specific to medical datasets, such as the Heart Disease dataset, have illustrated the effectiveness of hybrid approaches combining feature selection and dimensionality reduction for enhancing MLP performance.[46,47,48]

Through comparisons against baseline MLP models, this work also provides a richer picture of how the interplay between feature selection, dimensionality reduction, and neural network behaviour interacts with a variety of datasets.

One way to make your ML model better is to do feature selection which would help in reducing the dimension which might help to reduce overfitting and also interpretability of the model. Two well-known feature selection methods, Genetic Algorithm (GA) and Principal Component Analysis (PCA), are also investigated to optimize Multilayer Perceptron (MLP) networks in this study. These methods take different approaches and are useful in their own right for analyzing complex datasets.

Genetic Algorithm(GA): GA is a heuristic optimization technique that draws inspiration from the principles of natural selection and genetics. It proceeds by evolving a population of candidate solutions through generations, guided by selection, crossover, and mutation operations.[49] Specifically referring to feature selection, GA selects an optimal subset of features that improves the overall performance of the model by removing irrelevant or redundant attributes.[50]

Genetic Algorithms starts from an initial population of random feature subsets, and evaluates them based on a fitness function, usually the accuracy or some other performance metric of the model trained with them. The algorithm iteratively improves the population by mating leading sets through crossover and adding diversity through mutation. Such exploration makes GA as a method for searching through an enormous space of possible solutions.[51]

The GA is powerful since it can adapt to handle various datasets. As a case in point,[52] employed GA to select features in biomedical databases, leading to considerable enhancements in classification performance. GA has also been used to enhance an image dataset such as TinyFace where they successfully removed much noise and improved mmlp results due to GA comprehension of critical features of the dataset.[53]

Yet, the GA is computationally expensive, because this algorithm requires the training of machine learning models mentioned previously to be applied in the evaluation step. Regardless of that constraint, GA still is a strong tool for selecting features, especially for cases that early methods have a challenge reaching solutions.[54]

Principal Component Analysis (PCA) is a statistical method which aims at dimensionality reduction by converting the original features of a dataset into a new set of uncorrelated variables known as their principal components.[55] They are sorted so that the first few retain most of the information, and they are ordered by amount of variance captured in the data. PCA identifies the components that retain most of the variance, thereby decreasing the dimensionality of the dataset while maintaining its important structure.

This is beneficial in scenarios with high dimensional data, where a lot of features can be correlated and lead to poorer model performance. It is, however, a much more efficient way, utilizing techniques equivalent to linear algebra, e.g. eigenvalue decomposition or singular value decomposition, to find the principal components.[56]

PCA has been found effective in many areas, such as image recognition, where PCA is often used to preprocess large dataset before putting them through a neural network. For example, [25] used PCA to minimize a facial recognition dataset's feature space leading to reduced training times and better generalization. Measurement data in health care can also benefit from PCA for a more accessible data structure that improves prediction model readableness.[57]

Although PCA is a powerful tool, it has certain limitations. Being a linear method, it may not perform well with datasets that have non-linear relationships. Furthermore, PCA does not consider the relationship of features with the target variable explicitly; hence, it may yield a non-optimal closure for some tasks. Nevertheless, PCA continues to play an essential role in techniques for dimensionality reduction, especially once paired with other techniques for feature selection such as genetic algorithm (GA).[58]

Feature selection is a strength of both GA and PCA, making them complementary methods. Do note that GA can handle more complex and non-linear relationships and is more flexible. In contrast, PCA works exceptionally well at compressing high-dimension datasets into smaller, most useful datasets with minimal loss of meaningful data, providing computational efficiency. The ensemble of these techniques results in strong model feature set that improves models for different domains.

## METHOD

In this study, three datasets — TinyFace, Heart Disease, and Iris that span different domains and complexities are used. They also represent a solid basis for testing the effectiveness of feature selection and dimensionality reduction methods on MLP networks.

**TinyFace Dataset**

TinyFace: A Large-scale Benchmark for Low-resolution Face Recognition (LRFR)Research PaperThe TinyFace dataset is a large-scale benchmark dataset for research of low-resolution face recognition (LRFR). The dataset is composed of 5,139 identified identities, with 169,403 original low resolution (LR) facial images of 20×16 pixels average size. The dataset contains face images collected in uncontrolled environments, in which there is a great deal of variability, including pose variations, illumination conditions, occlusion, as well as background changes, which make it a particularly challenging dataset for a machine learning model.[59] Instead of down-sampling image datasets, different from previous works, TinyFace emphasizes on natively low-resolution images, which guarantees an evaluating framework for comparing LRFR techniques realistically.

This dataset has been used a lot in testing deep learning models as well as image preprocessing techniques.

The study uses TinyFace to investigate the connections of feature extraction, dimensionality reduction, and MLP network classification accuracy. The VGG19 architecture was employed for features extraction to accommodate the small image size and noise characteristics of the dataset.[60]

**Heart Disease Dataset**

In a 1988 publication by Janosi, Steinbrunn, Pfisterer, and Detrano, the Heart Disease dataset was created, serving as a classic benchmark in the medical diagnostics domain.[61] It has features of patients in relation to the presence of heart disease, with a total number of 303. They include 13 clinical and physiological features, including age, sex, chest pain type, resting blood pressure, serum cholesterol, fasting blood sugar, and more. These observations have a binary target variable, which puts patients with no heart disease into one category "Diameter" and patients with heart disease into another category "Narrowings".

It has garnered acceptance within the research community due to its small footprint and high-quality annotations, which are vital for deploying and validating machine learning models in the healthcare domain. In this study one dataset is used to test the effectiveness of feature selection methods such as that of genetic algorithms on MLP performance. The cleanliness, organization and sparsity (0,2 % missing values) of the data make it ideal for experimentation with classification accuracy and model optimization.

**Iris Dataset**

The Iris dataset is a basic dataset introduced by Fisher in 1936 and widely used for pattern recognition and machine learning. It has 150 samples among three classes of iris: Setosa, Versicolor, and Virginica.[62] All the four features (sepal length, sepal width, petal length, and petal width) are numeric and there are no NAN values. The labels classifies the flowers into three species.

Questions like those above are the reason why Iris dataset is the one of most common data set used for testing algorithms because it is simple and has a well defined structure. For example, in this study it is employed to analyze the effect of the feature selection and dimensionality reduction techniques (as PCA) on the classification accuracy of the MLP networks. The dataset is small in size and low in dimensionality, yet it serves as a good benchmark to analyze the performance of these techniques in optimizing the model.

**Experimentation and Data Processing Methods**

This segment details the experimentation and data processing methods employed to enhance the classification performance of a Multilayer Perceptron (MLP) network, especially on the difficult TinyFace dataset. Seventeen experiments were conducted, which included feature extraction, dimensionality reduction, and an iterative evaluation of model matrices to account for the complexities exhibited by the dataset. Well-being assessment based on 2020 face data

There were stepped design and implementation issues for the TinyFace dataset, most notably the low resolution of the face images (generally, 20×16 pixels resolution) and that there are plenty of classes, but only one or two samples available from each class. Such features require rigorous preprocessing and tuning to improve the classification rates of the MLP network.

**Extracting Features using VGG19**

Considering the resolution and noise in the TinyFace dataset, the use of the VGG19 convolutional neural network for feature extraction was justified. We applied VGG19 — a well-known deep learning model that excels at tasks related to images —to generate high-dimensional features of the original images. As the images of the dataset did not fit into the input size of VGG19, they were scaled to 224×224 pixels using the. PIL Image library's resize((224, 224)) function. This step of preprocessing helped make the image compatible with the VGG19 architecture, which made feature extraction possible.[63]

**PCA for Dimensionality Reduction**

PCA was used to reduce the high-dimensional feature space derived from VGG19, reducing the noise and redundant data impacts. Two settings of PCA (number of components (n_components) = 50 and 80) were explored. As expected, these arrangements obtained classification accuracies of 0,22 and 0,18, respectively, highlighting the extreme difficulties of dealing with the type of noisy and low-quality data present here.[64] This underscored that more scope exists for improvement in pre-processed data, as well as for model optimization.

**Resize Methods**

Due to the significance of image resizing for the subsequent feature extraction, the performance of different resizing methods for classification was analyzed. One option would be using Image. Resampling. Tried LANCZOS with some padding added. Nevertheless, the results (0,21 and 0,18, accuracy) suggested that using different resizing methods did not have a significant effect on the MLP network. This implied that the main difficulties were due to the dataset's own features and not related to the resizing algorithm itself.[65]

**Refining Dataset and Noise Reduction**

Due to the class imbalance and noise in the TinyFace dataset, preprocessing steps were introduced. To reduce the consequences of extreme sparsity, classes with fewer than five samples were discarded. This improved refinement decreased the variability in the dataset making the training more consistent. Following that, the newly refined dataset was employed in subsequent tests to assess how dimensionality sequence changes affected classification results.[66]

**Data Collection and Processing for Iterative Experimentation**

Iterative experimentation approaches were performed on the cleansed dataset, merging padded and non-padded resize methodologies, PCA with 50 and 100 segments, and default feature selection. 5-fold cross-validation showed best classification using padded, non-reduced features (CC = 0,477). These findings highlighted how retaining raw features was important for certain scenarios where data is both noisy and low resolution. Although data augmentation has not been employed in this step, it could be an effective way to further improvement, especially to overcome a few classes have limited sample size.[67,68,69]

**RESULTS AND DISCUSSION**

In this section, we present our mathematical findings from our structured experimental study on the optimization of Multilayer Perceptron (MLP) training networks with either feature selection using Genetic Algorithm (GA) and Principal Component Analysis (PCA) for dimensionality reduction. The evaluation includes three different datasets, namely TinyFace, Heart Disease, and Iris, and utilizes a number of controlled experiments to investigate the Effects of The Techniques On Classification Performance.

**TinyFace Dataset**

Besides, the TinyFace dataset was a challenge because of the low-resolution, noise, and class imbalance. The experiments yielded the following results:

1. MLP with Default Parameters: Default configuration for MLP training resulted in only 0,478 accuracy, very low considering how our data is noisy and low resolution, thus this training is not showing optimal feature learning.
2. GA Feature Selection MLP: This was applied with GA for feature selection improved up to 0,545 in accuracy. This highlights GA's potential for identifying and keeping the most significant features for classification, even in difficult datasets.
3. PCA Dimensionality Reduction: After obtaining 25,000 features, PCA was applied to filter the features down to 1,400, retaining 95 % of variance. However, the accuracy decreased to 0,125 which can be expected as dimensionality reduction can result in information loss especially when working with highly sparse and noisy datasets.

**Heart Disease Dataset**

The Heart Disease dataset is a matrix of structured, well-labeled clinical data, where trends in performance were as follows:

1. MLP with Default Parameters: We hedged on the accuracy on a very high scale of 0,85 as to show the relevance of the dataset to be used with machine learning tasks.
2. GA Feature Selection MLP: Adding in GA raised that to 0,90, highlighting how feature selection can significantly improve classification performance by removing highly correlated, contributing less significant features.
3. PCA Dimensionality Reduction: We applied PCA to reduce the number of features to 2 of them PG and TCCA almost retaining 95 % of the variance. This resulted in an accuracy of 0,68, giving insights about the dimensionality reduction and accuracy of models with interdependent features.

**Iris Dataset**

Since the Iris dataset is simple and low-dimensional, it was used as a baseline to evaluate model performance in the best-case scenario:

1. MLP with Default Parameters: As a result, the MLP achieved a 1,0 accuracy indicating that the dataset was a good candidate for use with standard machine learning algorithms.
2. GA Feature Selection MLP: GA based Feature selection reduced the feature set but retained the perfect accuracy of 1,0 which validates the performance of GA based Feature Selection as the optimal sub features selected provide perfect performance without loosing on performance.
3. PCA Dimensionality Reduction: Thus, PCA reduced the no. of features further to 1 subject to

retaining 95 % of variance and gave an accuracy level as high as 0,96. This outcome highlights PCA's capabilities for low-dimensional and noise lee datasets.

**Comparative Analysis**
The comparison between results across data sets indicates a pattern of differences:
- GA Feature Selection improved MLP for all datasets, with the greatest differences observed for the TinyFace and Heart Disease datasets. But its capacity to detect important features and discard superfluous data was remarkably useful for noisy or high-dimensional problems.
- Training PCA Dimensionality Reduction on a dataset caused PCA to have a very tight performance iteration on the dataset (e.g., Iris), while performance was poor in complex and noisy datasets (such as, TinyFace). The substantial loss in accuracy in such cases indicates that the linear nature of PCA might not be able to model nonlinear relationships between features or cope with sparsity well.

The graphical representation provided in figure 1 acts as a visual tool to accompany the written explanation, allowing readers a quick and holistic understanding of the diverse outputs the Multilayer Perceptron (MLP) networks display across the treatment types in our experimental setup.

Data presented in figures therein serve as a pictorial representation which increases the interpretability and thus, accessibility of the empirical findings, and hence augment the academic narrative articulated in this results section.

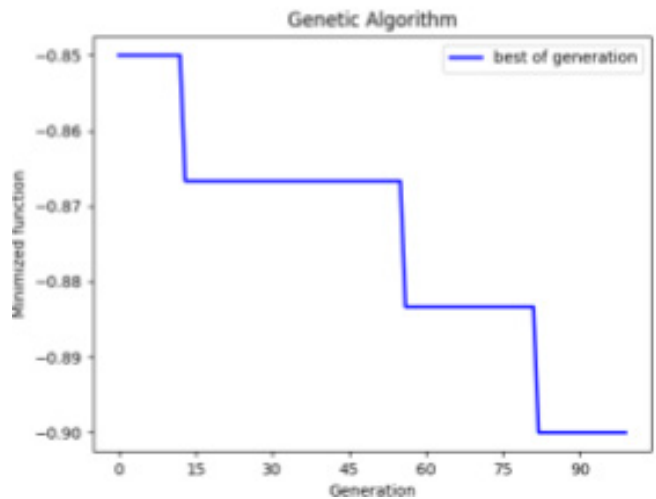

**Figure 1.** Progression of each generation, converging toward higher accuracy

The performance results are summarized in table 1, providing a concise overview of the outcomes across different configurations.

| Table 1. Comparative Training Accuracy of Multilayer Perceptron (MLP) Configurations across Datasets | | |
|---|---|---|
| Dataset | Model Configuration | Training Accuracy |
| tinyFace | MLP with default params | 0,478 |
| | GA Feature Selection | 0,545 |
| | PCA 95 reduced to 1400 features then trained on default | 0,125 |
| Heart Disease | MLP with default params | 0,85 |
| | GA Feature Selection | 0,90 |
| | PCA 95 reduced to 2 features then trained on default | 0,68 |
| Iris | MLP with default params | 1,0 |
| | GA Feature Selection | 1,0 |
| | PCA 95 reduced to 1 features then trained on default | 0,96 |

**CONCLUSION**
In this study a thorough analysis of MLP network optimization via dimensionality reduction and feature selection is given. Utilizing GA and PCA across three dramatically dual datasets of differing intricacy, the findings granted expose to the complex routine of ameliorating model performances.

The results proved that GA consistently enhances accuracy by choosing optimal subsets of features, especially in datasets with higher dimensionality or noise, like TinyFace and Heart Disease. In contrast, PCA started to show good results in the low-dimensional and structured datasets such as Iris where the usage of the PCA decreased the computation complexity without noticeably affecting the accuracy. This showed that GA and PCA had different strengths complementary to each other, suggesting that no optimal strategy could be devised at a general level and that such strategies would have to be dataset-specific.

The study highlights the nuanced trade-offs between feature selection and dimensionality reduction as well. GA is proficient in identifying non-linear relationships and identifying optimal subsets of features, whereas PCA is efficient in terms of computational cost for obtaining low-dimensionality representations of data that exhibit linear correlations among features.

As problems become more complex, machine learning can take this work as motivation for the need of problem-specific optimization. Lastly, future research might look into hybrid methods, where algorithmic approaches are interleaved to work on real-life, large-scale datasets with adaptive optimization methods. Indeed, this trend shows promise in improving on the efficiency and generalization of neural networks across many fields, from health care to computer vision.

Technologies and Applications (eSmarTA) [Internet]. IEEE; 2023. p. 1–8. Available from: https://ieeexplore.ieee.org/document/10293415/

**FINANCING**

This work is supported from Jadara University under grant number [Jadara-SR-Full2023].

**CONFLICT OF INTEREST**

The authors declare that the research was conducted without any commercial or financial relationships that could be construed as a potential conflict of interest.

**AUTHORSHIP CONTRIBUTION**

*Conceptualization:* Mohammad Subhi Al-Batah, Mowafaq Salem Alzboon, Muhyeeddin Alqaraleh.
*Data curation:* Mohammad Subhi Al-Batah, Mowafaq Salem Alzboon, Muhyeeddin Alqaraleh.
*Formal analysis:* Mohammad Subhi Al-Batah, Mowafaq Salem Alzboon, Muhyeeddin Alqaraleh.
*Research:* Mohammad Subhi Al-Batah, Mowafaq Salem Alzboon, Muhyeeddin Alqaraleh.
*Methodology:* Mohammad Subhi Al-Batah, Mowafaq Salem Alzboon, Muhyeeddin Alqaraleh.
*Project management:* Mohammad Subhi Al-Batah, Mowafaq Salem Alzboon, Muhyeeddin Alqaraleh.
*Resources:* Mohammad Subhi Al-Batah, Mowafaq Salem Alzboon, Muhyeeddin Alqaraleh.
*Software:* Mohammad Subhi Al-Batah, Mowafaq Salem Alzboon, Muhyeeddin Alqaraleh.
*Supervision:* Mohammad Subhi Al-Batah, Mowafaq Salem Alzboon, Muhyeeddin Alqaraleh.
*Validation:* Mohammad Subhi Al-Batah, Mowafaq Salem Alzboon, Muhyeeddin Alqaraleh.
*Visualization:* Mohammad Subhi Al-Batah, Mowafaq Salem Alzboon, Muhyeeddin Alqaraleh.
*Writing – original draft:* Mohammad Subhi Al-Batah, Mowafaq Salem Alzboon, Muhyeeddin Alqaraleh.
*Writing – revision and editing:* Mohammad Subhi Al-Batah, Mowafaq Salem Alzboon, Muhyeeddin Alqaraleh.